%% file: main.tex
\documentclass[letterpaper, 10 pt, conference]{ieeeconf}  

\IEEEoverridecommandlockouts                              

\title{\LARGE \bf
Privacy Risks in Reinforcement Learning for Household Robots
}

\author{Miao Li$^{1}$, Wenhao Ding$^{1}$, Ding Zhao$^{1}$
\thanks{$^1$Carnegie Mellon University, Pittsburgh, PA 15213, USA.
Email: \{limiao, wenhaod, dingzhao\}@andrew.cmu.edu}%
}

\usepackage{cite}
\usepackage{amsmath}
\usepackage{graphicx}
\usepackage{amsfonts}
\usepackage{color}
\usepackage{subfigure}
\usepackage{subcaption}
\usepackage{multirow}
\usepackage{booktabs}
\usepackage{cite}
\usepackage{xcolor}
\usepackage{hyperref}

\begin{document}

\maketitle
\thispagestyle{empty}
\pagestyle{empty}

\newpage


\begin{abstract}
The prominence of embodied Artificial Intelligence (AI), which empowers robots to navigate, perceive, and engage within virtual environments, has attracted significant attention, owing to the remarkable advances in computer vision and large language models.
Privacy emerges as a pivotal concern within the realm of embodied AI, as the robot accesses substantial personal information.
However, the issue of privacy leakage in embodied AI tasks, particularly concerning reinforcement learning algorithms, has not received adequate consideration in research.
This paper aims to address this gap by proposing an attack on the training process of the value-based algorithm and the gradient-based algorithm, utilizing gradient inversion to reconstruct states, actions, and supervisory signals.
The choice of using gradients for the attack is motivated by the fact that commonly employed federated learning techniques solely utilize gradients computed based on private user data to optimize models, without storing or transmitting the data to public servers. Nevertheless, these gradients contain sufficient information to potentially expose private data.
To validate our approach, we conducted experiments on the AI2THOR simulator and evaluated our algorithm on active perception, a prevalent task in embodied AI. 
The experimental results demonstrate the effectiveness of our method in successfully reconstructing all information from the data in 120 room layouts. Check our \href{https://rl-privacy.github.io/RL-gradient-inversion/}{website} for videos.
\end{abstract}

\section{INTRODUCTION}

\input{introduction}

\section{RELATED WORK}
\input{relatedwork}

\section{PRELIMINARY}
\input{preliminary}

\section{METHOD}
\input{method}

\section{EXPERIMENT}

\input{experiment}

\section{CONCLUSION}
\input{conclusion}

\bibliographystyle{IEEEtran}

\input{main.bbl}
\end{document}

%% file: introduction.tex
The advent of recent large models has brought about remarkable achievements in human-like dialog generation~\cite{ouyang2022training} and controllable image generation~\cite{ramesh2022hierarchical}. These achievements highlight the promising potential of leveraging artificial intelligence (AI) to enhance human experiences and to establish the foundation for embodied AI.
The next significant milestone in the advancement of general AI revolves around the exploration of embodied systems~\cite{duan2022survey} that possess the ability to navigate, perceive, engage, and successfully tackle tasks within the physical world~\cite{brohan2023rt}.

The process of collecting embodied datasets presents more privacy concerns compared to language and vision tasks, since the robot operates within real-world environments, executing policies, and gathering observational signals. These environments often contain personal information \cite{sakuma2008privacy, wang2019functionnoiseQ}, which presents challenges during data collection and model training.
\textit{Federated Learning} framework (FL)~\cite{zhang2021survey} is introduced to enable robots deployed in the private environment to conduct online training without distributing any data.
This framework allows local storage of private data on individual machines, with only the gradients, calculated based on private information within the environment \cite{mcmahan2017fedavg, konevcny2015federated, konevcny2016federated, reddi2020adaptivefedopt}, being transmitted to the central server. These gradients, which originate from multiple private servers, are then aggregated to update the model on the central server. The updated model is then sent back to the private servers for the next iteration. As a result, the model and gradients are accessible to potential adversaries on the central server, whereas the data remains accessible only to private servers.

However, relying solely on the transmission of gradients in FL still leaves room for vulnerability to gradient inversion techniques~\cite{zhu2019deepleakage,geiping2020inverting,hatamizadeh2022gradvit,zhao2020idlg}, a type of method that can reconstruct input images and labels from the corresponding gradients in classification tasks.
Gradient inversion algorithms search for synthetic images in latent spaces using generative networks \cite{jeon2021gradientgenerative} or directly optimize a synthetic image in the image domain, initialized by heuristic methods. 
The reconstruction loss compares the real gradient sent by the private server with the gradient produced using the synthetic data. The visual quality of the reconstructed image is ensured by loss functions based on prior knowledge, e.g., smoothness of natural images~\cite{zhu2019deepleakage, 2019userlevel, geiping2020inverting, hatamizadeh2022gradvit}.
These techniques can reconstruct high-resolution images with recognizable patterns using gradients from mini-batches with small batch sizes \cite{geiping2020inverting, yin2021seethroughgrad}.
Although these studies cannot make all images recognizable, they highlight the risk of privacy, that is, the leakage of private training data associated with FL.

\begin{figure}
    \centering
    \includegraphics[width=0.47\textwidth]{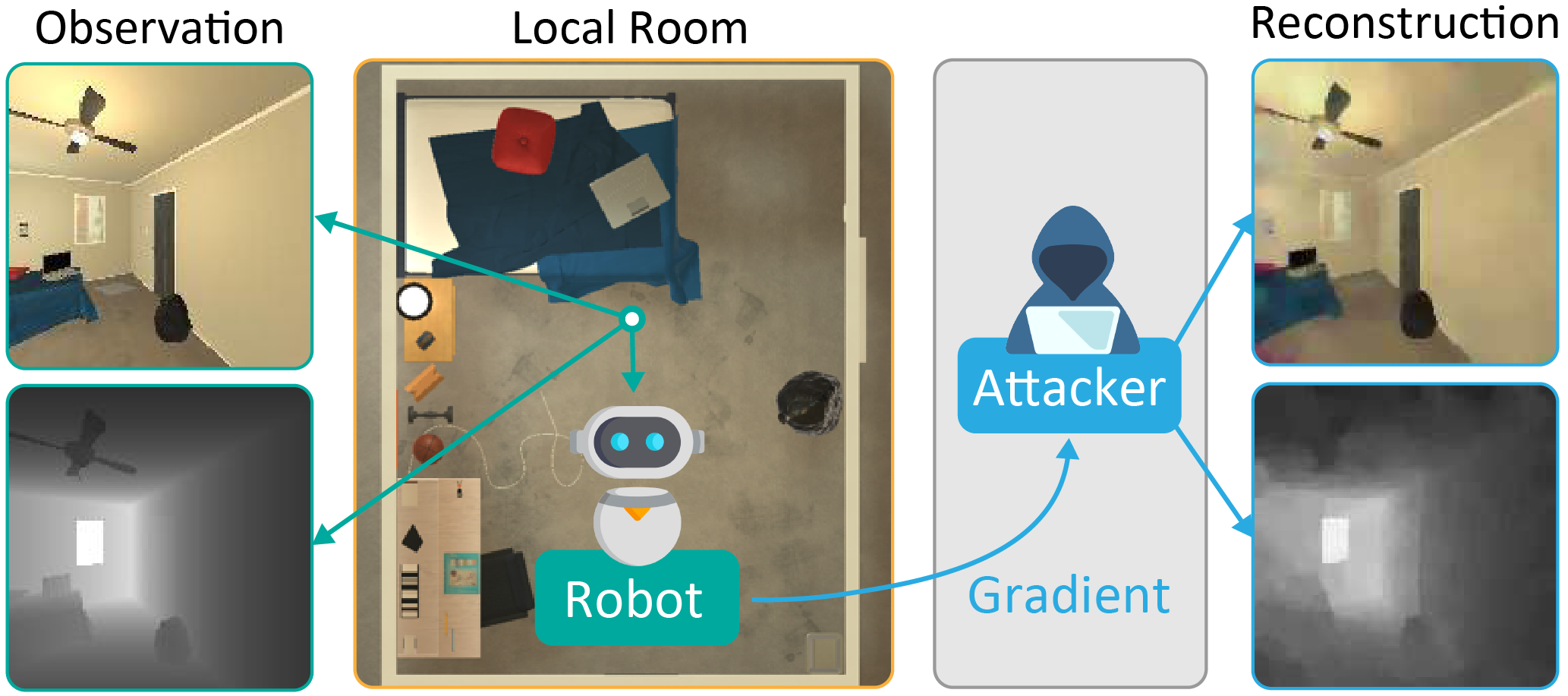}
    \caption{Our method (Attacker) can reconstruct the images of your untidy room and window orientation from the gradient.} 
    \label{fig:motivation}
    \vspace{-7mm}
\end{figure}


Motivated by the existing literature on gradient inversion in classification tasks, our research aims to investigate the privacy leakage issue in the context of embodied AI and specifically focuses on the gradient inversion attack to widely used reinforcement learning (RL) algorithms, an area that has received limited attention so far. 
Fig.~\ref{fig:motivation} provides an illustrative example of this problem, demonstrating how an attacker can successfully reconstruct RGB and depth images of a private room solely from the gradient.
Unlike classification tasks where the input typically consists of a single RGB image, applying gradient inversion techniques to RL poses unique challenges due to the multimodal nature of the input information including various images and vector states.
Our objective is to reconstruct all these inputs exclusively from the gradients 
shared by the RL algorithm.
The most relevant work to this topic is \cite{pan2019howyouact}, which reconstructs the location map from the model parameter of the Deep Q Network (DQN)~\cite{mnih2015dqn}. However, this approach is limited in that it supports only one map for one model and focuses on a trained model other than the learning process.

In this work, we introduce two novel gradient inversion approaches targeting the value-based RL algorithm and the gradient-based RL algorithm, namely \textit{Deep Q-learning Gradient Inversion} (\textbf{QGI}) and \textit{REINFORCE Gradient Inversion} (\textbf{RGI}). 
We address the task of reconstructing multimodal data from the gradient of RL algorithms.
Specifically, we initialize the reconstructed state with Gaussian noises and optimize the cosine similarity between the gradients of the true and reconstructed data. 
Compared to image classification, gradient inversion in RL is challenging due to multimodal input and complex objective functions. 
Thus, we propose a pipeline with rule-based methods for action and supervisory signals and optimization-based methods for multimodal input based on objective-agnostic gradient calculation.
We follow the white-box assumption~\cite{hitaj2017GANmodelinversion} and the honest-but-curious assumption~\cite{geiping2020inverting} of the potential adversary. 
We evaluate QGI and RGI in the AI2THOR~\cite{kolve2017ai2thor} simulator on the active perception task~\cite{ding2023learning, kotar2022interactron, ammirato2017dataset}, a popular task in embodied AI to navigate agents to obtain a higher detection accuracy of the given object. 
Since it requires multimodal information about the private user room, we select it as an example task to apply the gradient inversion attack.
Our main contributions are summarized below:
\begin{itemize}
    \item 
     As far as we know, QGI and RGI are the first frameworks to investigate the privacy leaking problem by reconstructing data from the gradient in RL.  Based on objective-agnostic gradient calculation, the proposed methods are generalizable to other RL algorithms.
    \item We provide novel pipelines to reconstruct multimodal information on states, actions, and supervision signals for value-based and policy gradient-based algorithms. 
    \item We evaluate our method in a realistic simulation of the active perception task and show that our QGI and RGI can successfully reconstruct all information. 
\end{itemize}

%% file: relatedwork.tex
\textbf{Privacy attack in Deep Learning.}
Private training data is at risk of attacks, including membership inference, model inversion, and gradient inversion \cite{rigaki2020survey}.
Membership inference assumes that the adversary has access to certain data and attempts to verify whether the data is used in the training procedure \cite{shokri2017membership, carlini2021extracting}, while in robotic tasks, private environments tend to stay inaccessible to adversaries, for example, bedrooms of private property.
In contrast, model inversion and gradient inversion assume that private data is inaccessible and the adversary attempts to reconstruct the data, which are of higher significance for robotic tasks.
While model inversion reveals the training data from the trained model \cite{an2022mirror, mordvintsev2015deepdream, hitaj2017GANmodelinversion, yin2020deepinversion}, gradient inversion tackles the reconstruction from the gradients shared in federated, collaborative, decentralized, and distributed learning \cite{zhu2019deepleakage, geiping2020inverting, yin2021seethroughgrad, jeon2021gradientgenerative, jin2021cafe, hatamizadeh2022gradvit}. 

\textbf{Gradient inversion.}
Previous work in gradient inversion focused mainly on image classification tasks.
\cite{zhu2019deepleakage} first formulated the gradient inversion technique and proposed an end-to-end method DLG to reconstruct both images and labels for the classification task from single-sample gradients and batched gradients. 
\cite{zhao2020idlg} proposed an efficient method to identify the one-hot label from a single-sample gradient.
\cite{geiping2020inverting} extended gradient inversion to high-resolution images by minimizing a prior loss proven effective in image denoising \cite{rudin1992denoise} and model inversion \cite{mordvintsev2015deepdream} and a magnitude-agnostic loss function of cosine similarity. We adopt this loss function in this work.
\cite{yin2021seethroughgrad} increased the accuracy in large batch sizes by leveraging prior loss functions and penalizing differences in the results of multiple trials. 
To achieve better image reconstruction, generative networks are considered in~\cite{2019userlevel, jeon2021gradientgenerative}.
Gradient inversion is also investigated in language tasks~\cite{zhu2019deepleakage, deng2021tag}.
To the best of our knowledge, gradient inversion is unexplored in the realm of multi-modal robotic tasks or reinforcement learning.

\textbf{Privacy in Reinforcement Learning.}
Privacy has been studied in existing Reinforcement Learning literature for a long time.
However, most works dedicated to privacy-preserving \cite{sakuma2008privacy, vietri2020privatePAC&regret, wang2019functionnoiseQ, garcelon2021localDP, taherisadr2023adaparl} other than revealing the risk of privacy leakage.
The most relevant work for this paper in Reinforcement Learning is \cite{pan2019howyouact}. 
They reconstructed the training environment from the trained policies, demonstrating the privacy risk of releasing trained models.
Based on the prior knowledge, they used the genetic algorithm to search for the configuration of the environment.
Although they studied the sensitive information contained in trained models, the risk associated with the training procedure remains ambiguous.
In this work, we investigate the possibility of revealing private state, action, and Q-values from gradients. 

%% file: preliminary.tex
Prior to presenting the proposed gradient inversion approaches aimed at RL algorithms, 
this section outline the formulation of the gradient inversion problem and the two victim algorithms.

\textbf{Gradient Inversion.}
Gradient inversion aims to reconstruct training data from the gradient of the learnable model~\cite{zhu2019deepleakage} calculated by backpropagation~\cite{BP}, which can be used by the potential adversary when the training data is invisible but the gradient is shared, for example, in FL.
Given the input $x$, the supervisory signal $u$ (e.g., label in classification), the model $F$ with parameters $w$, the output $F(x;w)$, and the objective function $J$, then the gradient is $g=G(x,u)=\nabla_w J(F(x;w), u) = \frac{\partial J(F(x;w),u)}{\partial F(x;w)} \frac{\partial F(x;w)}{\partial w}$, where $G$ represents gradient calculation.
The gradient carries the information of $x$ in $\frac{\partial F(x;w)}{\partial w}$ and the information of $u$ in the $\frac{\partial J(y,u)}{\partial y}$. 
The goal of gradient inversion is to reconstruct $(x,u)$ by generating $(x^{\text{rec}}, u^{\text{rec}})$ that produces the gradient $g^{\text{rec}}$, which is sufficiently close to the true gradient $g$. 
Assuming that the adversary has white-box access to the model $F$ and the objective $J$, the reconstructed gradient is $g^{\text{rec}}=G(x^{\text{rec}},u^{\text{rec}})=\nabla_w J(F(x^{\text{rec}};w), u^{\text{rec}})$.
The adversary minimizes the gradient matching loss function $L(g^{\text{rec}},g)$ to obtain the reconstruction $x^{\text{rec}}, u^{\text{rec}}$ as follows:
\begin{align}
    \arg\min_{x^{\text{rec}}, u^{\text{rec}}} L(G(x^{\text{rec}}, u^{\text{rec}}),g).
    \label{eq:key of gradient inversion}
\end{align}
Optimization-based gradient inversion uses backpropagation to obtain gradients of the reconstruction based on $L(g^{\text{rec}},g)$, while rule-based gradient inversion solves (\ref{eq:key of gradient inversion}) directly.

\textbf{Markov Decision Process, DQN and REINFORCE.}
We consider the Markov Decision Process (MDP) as the mathematical framework to model decision-making problems in reinforcement learning.
MDP consists of the state space $s\in \mathcal{S}$, action space $a\in \mathcal{A}$, reward function $r\in \mathcal{R}$, and a transition model $p\in \mathcal{P}$. The future discounted return at timestep $t$ is $R_t = \sum_{t'=t}^T \gamma^{t'-t}r_{t'}$, where $\gamma$ is the discount factor and $T$ is the maximal step. $\pi$ is a policy mapping states to actions.
In this paper, we investigate the gradient inversion attack on DQN~\cite{mnih2015dqn}, which estimates the optimal action-state value function (Q-value) $Q^*(s, a)$.
Empirically, this $Q^*(s, a)$ is parameterized with neural networks and the objective is to minimize the difference between the predicted Q-value $\hat{Q}(s,a)$ of the selected action $a$ and the target Q-value $Q(s,a)$ with transitions $(s_t, a_t, s_{t+1}, r_t)$. 
In this work, we adopt the mean square error (MSE) as the objective.
We also investigated the gradient inversion attack for the REINFORCE algorithm~\cite{reinforce} with the entropy term, which utilizes the policy gradient to update the policy. 
The gradient of the policy model is estimated as
\begin{align}
    \nabla_{\theta} J = \mathbb{E}[\log \pi_{\theta}(a|s) R_t] - \alpha \nabla_{\theta} \pi_{\theta}(s) \log \pi_{\theta}(s),
    \label{eq: reinforce objective}
\end{align}
where $J$ is the total objective and $\alpha$ is the weight of entropy.


%% file: method.tex
In this section, we introduce gradient inversion in reinforcement learning, a method to reconstruct the training data sample from the gradient of RL algorithms, including the multimodal state $s$, action $a$, and supervisory signal, for example, the target Q-value $Q(s, a)$ or the return $R_t$. 
Note that the reward may not be recoverable when not used in the gradient calculation. 
The input data are reconstructed by optimization-based gradient inversion. To reconstruct the input of different types of models, we propose separate optimization stages. 
To cope with complex objective functions of the RL algorithms, we propose to calculate the reconstructed gradient $g^{\text{rec}}$ from the error direction, instead of the original objective function. 
With the objective-function-agnostic gradient calculation, our method can be transferred to other RL algorithms with various objectives.
Action and supervision signals are reconstructed using rule-based methods.

We demonstrate our method with the active perception task, where the state $s$ contains the depth image $s_d$, RGB image $s_i$, and the coordinate $s_c$ that indicates the position of the target object in the images.
The images are processed by convolution layers, while the coordinate vector is processed by linear layers. 
The features are then concatenated and processed with linear layers.
The action space is discrete with seven movement options.
The reward $r$ is defined by the confidence score obtained from an object detection model.

\begin{figure*}
    \centering
    \includegraphics[width=\textwidth]{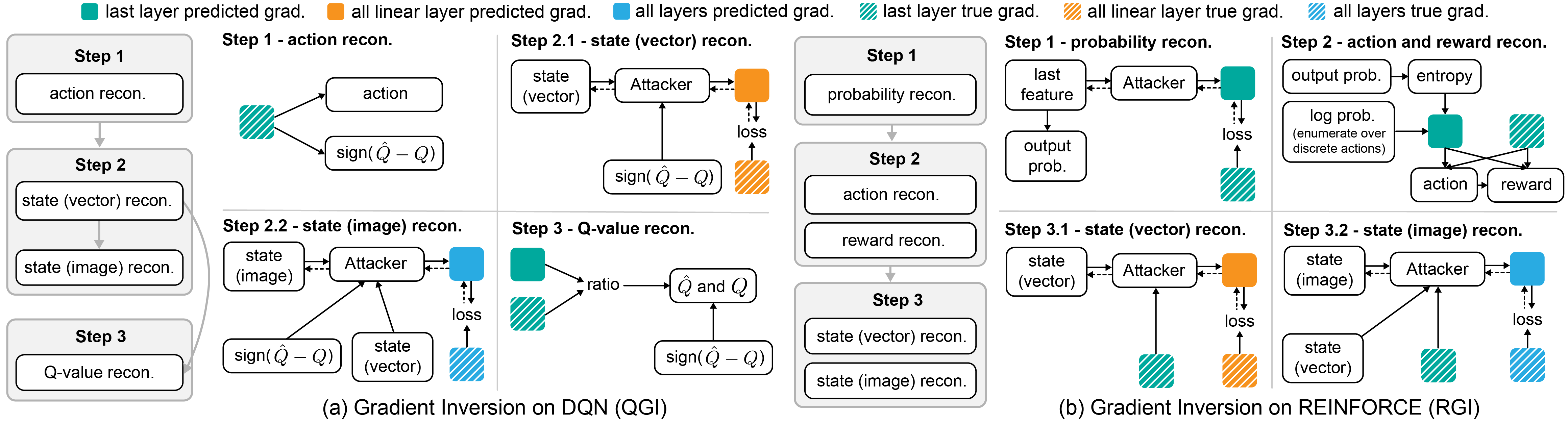}
    \caption{Pipeline of the proposed gradient inversion attack on DQN (left) and REINFORCE (right). The left part of each image shows the steps to reconstruct the multimodal information, and the right part explains each step in detail.}
    \label{fig:pipeline}
    \vspace{-3mm}
\end{figure*}

We focus on the value-based and policy-gradient-based RL algorithms and 
illustrate details of the proposed method through the Gradient Inversion attack on DQN \cite{mnih2015dqn} (QGI) and REINFORCE \cite{reinforce} (RGI). 
For conciseness, we use $\hat{Q}$ and $Q$ as shorthand representations of $\hat{Q}(s,a)$ and $Q(s,a)$.
The pipelines for QGI and RGI are shown in Fig.~\ref{fig:pipeline}.
QGI contains 3 steps. 
In {\bf step 1}, the action is identified using a rule-based method.
In {\bf step 2}, we reconstruct the multimodal state by optimization in two stages. Stage 1 targets the input of linear layers, including the vector state $s_c$. Stage 2 reconstructs the image states, namely $s_i$ and $s_d$. 
In {\bf step 3}, we estimate the predicted Q-value $\hat{Q}^{\text{rec}}$ by feeding the reconstruction and the target Q-value $Q^{\text{rec}}$ using a rule-based method. 
RGI also contains 3 steps.
In {\bf step 1}, we reconstruct the probability for each action by optimization-based gradient inversion with objective-agnostic gradient calculation.
In {\bf step 2}, we reconstruct the action and the return $R_t$ using a rule-based method.
In {\bf step 3}, the multimodal state is reconstructed in the same two stages as QGI. 
QGI can attack other algorithms that minimize element-wise error, while RGI is generalizable for policy-gradient-based algorithms.


\subsection{Optimization-based Multimodal State Reconstruction}
Optimization-based gradient inversion minimizes the gradient matching loss $L(g^{\text{rec}},g)$ by applying gradient descent to the reconstructed data $(x^{\text{rec}}, u^{\text{rec}})$
in (\ref{eq:key of gradient inversion}).
As shown in Fig.~\ref{fig:pipeline}, we separate state reconstruction into two stages for both QGI and RGI.
We first reconstruct the vector state $s_c$ and then reconstruct the image state $s_i$ in \textbf{stage 1} and $s_d$ with the vector state fixed in \textbf{stage 2}. 
In each stage, we initialize the input reconstruction with Gaussian noise and use gradient descent to update the reconstruction.
We optimize the reconstructed states with gradient matching loss $L(g^{\text{rec}},g)$, implemented by cosine similarity~\cite{geiping2020inverting}:
\begin{align}
    L(g^{\text{rec}},g) = 1 - \frac{\langle g^{\text{rec}},g \rangle}{\|g^{\text{rec}}\|_2\|g\|_2}.
\label{equ:gradient_matching}
\end{align}
For the image state, we add a total variation (TV) loss~\cite{rudin1992denoise, hatamizadeh2022gradvit, geiping2020inverting} as a prior term in addition to $L(g^{\text{rec}},g)$ to penalize noisy patterns, leading to the total loss $L_{\text{img}}=L(g^{\text{rec}},g) + \lambda( \text{TV}(s_i) + \text{TV}(s_d))$, where $\lambda$ is a weight. 

\subsection{Objective-Agnostic Gradient Calculation}
In the optimization-based gradient inversion, $g^{\text{rec}}$ is usually calculated by backpropagation after feeding the reconstructed training data, i.e. $g^{\text{rec}}=\nabla_w J(F(x^{\text{rec}};w), u^{\text{rec}})$.
As objective functions in RL may contain special calculations and data flow that compromise attack performance, we propose an objective-agnostic gradient calculation method. We interpret the gradient as the result of backpropagation from the error direction $\frac{\partial J}{\partial F(x;w)}$ other than the original objective $J$. 
When the batch size is 1 and the last layer of the network is a linear layer with a bias parameter, this error direction $\frac{\partial J}{\partial F(x;w)}$ is equal to the gradient of the bias parameter of the last layer $\nabla_{b_{\text{last}}}J(F(x;w),u)$. Thus, we propose a surrogate objective:
\begin{align}
    \tilde{J}(F(x^{\text{rec}}),u^{\text{rec}})=\langle \nabla_{b_{\text{last}}}J(F(x),u), F(x^{\text{rec}}) \rangle,
\end{align} 
and estimate $g^{\text{rec}}$ as $\tilde{g}^{\text{rec}} = \nabla_w \tilde{J} = \frac{\partial J}{\partial F(x;w)} \frac{\partial F(x^{\text{rec}};w)}{\partial w}$.
Given $g=\frac{\partial J}{\partial F(x;w)} \frac{\partial F(x;w)}{\partial w}$, $\tilde{g}^{\text{rec}}$ from $\tilde{J}$ is the same as $g$ from $J$ when the reconstructed data $x^{\text{rec}}$ is the same as $x$. 
Thus, we use $L(\tilde{g}^{\text{rec}},g)$ to reconstruct the input, namely the multimodal state. This surrogate objective allows for objective-agnostic gradient calculation and, therefore, our method can easily apply to other RL approaches with diverse objectives.

\subsection{Rule-based Reconstruction for QGI}
\subsubsection{Action Reconstruction}
Since the action space of DQN is discrete, 
we propose to directly identify the action from the gradient when the batch size is 1. 
Observe that the predicted Q-value vector $\hat{Q}(s)$ is obtained from the last linear layer with weight $W_{\text{last}}$ and bias $b_{\text{last}}$, 
and the objective $J(\hat{Q}(s,a), Q(s,a))$ extracts elements of $\hat{Q}(s,\cdot)$ and $Q(s,\cdot)$ that correspond to $a$ and only provides nonzero value for $\hat{Q}(s,a)$. 
$a$ can be obtained from the bias gradient 
\begin{align}
    \nabla_{b_{\text{last}}}J = \frac{\partial J(\hat{Q}(s,a),Q(s,a))}{\partial \hat{Q}(s,a)} 
    \underbrace{\frac{\partial \hat{Q}(s,a)}{\partial \hat{Q}(s,\cdot)}}_{\text{one-hot matrix}}
    \underbrace{\frac{\partial \hat{Q}(s,\cdot)}{\partial b_{\text{last}}}}_{\textbf{=\textbf{1}}}.
\end{align}
When $\frac{\partial J(\hat{Q},Q)}{\partial \hat{Q}}\neq 0$ and batch size is 1, $\nabla_{b_{\text{last}}}J$ is a one-hot vector, uniquely identifying the action $a$. 
When the batch size is larger, the bias gradient may not be one-hot. In this case, one solution is to enumerate the action combination and select the one with minimal gradient matching loss. 

\subsubsection{Q-value Reconstruction}
Since rewards are not used directly in gradient calculation, it is essential to first reconstruct the target Q-value and then utilize the target Q-value to estimate the reward.
Reconstruction of the target Q-value $Q^{rec}$ is implemented in 3 stages, (1) identifying the sign of the error $\Vec{n}=\text{sign}(\hat{Q}-Q)$ based on $\nabla_{b_{\text{last}}}$, (2) reconstructing the magnitude of the error $\|\hat{Q}-Q\|$, (3) reconstructing $Q$ based on the reconstructed error and the reconstruction of the predicted Q-value $\hat{Q}^{\text{rec}}$. 
With the MSE objective function $J(\hat{Q},Q)=(\hat{Q}-Q)^2$, the gradient is $g=\nabla_w J= \frac{\partial J(\hat{Q},Q)}{\partial \hat{Q}} \frac{\partial \hat{Q}}{\partial w} = 2(\hat{Q}-Q)\frac{\partial \hat{Q}}{\partial w}$.
Consequently, we obtain
$\frac{\|g\|}{\|g^{\text{rec}}\|} = 
\frac{\left\| \hat{Q}-Q \right\|}{\left\| \hat{Q}^{\text{rec}} - \tilde{Q} \right\|} 
    \frac{\|{\partial \hat{Q}}/{\partial w}\|}{\|{\partial \hat{Q}^{\text{rec}}}/{\partial w}\|}$,
where $\tilde{Q}$ is a constant used as the target Q-value during gradient inversion that produces the accurate error direction. 
Once the input data of the linear layers are obtained in step 2.1, we feed the reconstruction to the linear layers and obtain the reconstruction of the predicted Q-value $\hat{Q}^{\text{rec}}$. 
Since the reconstructions of linear layer inputs are usually accurate, 
$\frac{\|{\partial \hat{Q}}/{\partial w}\|}{\|{\partial \hat{Q}^{\text{rec}}}/{\partial w}\|}\approx1$.
The magnitude of the error between the predicted and the target Q-value is reconstructed by $\| \hat{Q}-Q \|^{\text{rec}} \approx \| \hat{Q}^{\text{rec}} - \tilde{Q} \| \frac{\|g\|}{\|g^{\text{rec}}\|}$.
Then target Q-value is reconstructed as $Q^{\text{rec}} \approx \hat{Q}^{\text{rec}} - \Vec{n} \| \hat{Q}-Q \|^{\text{rec}}.$
When the batch size is larger than 1,
the target Q-value could be reconstructed by solving linear equations between the error magnitudes and the gradient \cite{zhu2020r-gap, cocktail}.

\begin{figure*}[t]
    \centering
    \includegraphics[width=\textwidth]{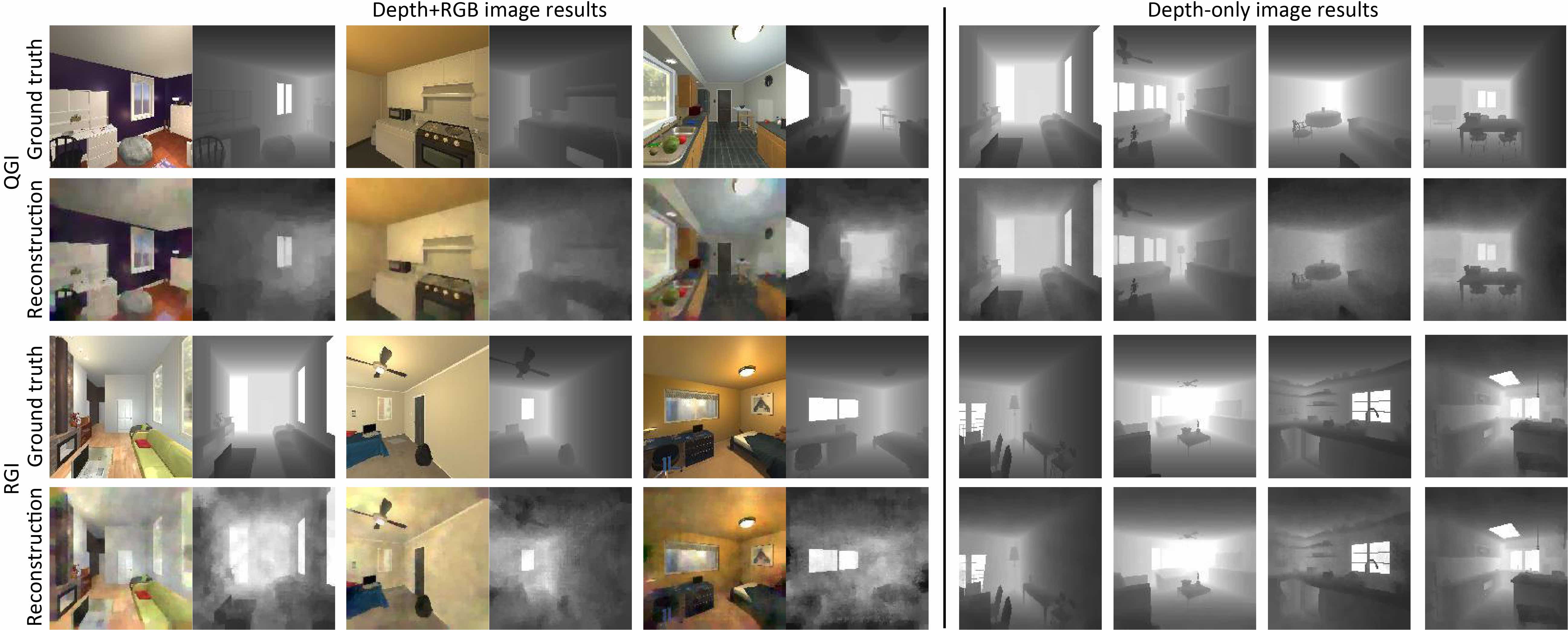}
    \caption{Qualitative results. The left shows the results of depth+RGB images (S1). The right shows the results of depth images (S2). The upper shows the results of QGI. The lower shows the results of RGI.}
    \label{fig:qualitative}
    \vspace{-5mm}
\end{figure*}

\subsection{Rule-based Reconstruction for RGI}
The action and return of REINFORCE are reconstructed by rule-based methods. 
After conducting optimization-based gradient inversion on the last linear layer to reconstruct the input using the surrogate objective, we feed the input reconstruction through the model and reconstruct the output, namely the probability of all actions as $p^{\text{rec}}$. 
The objective function of REINFORCE has two components, namely the action-and-return-dependent policy gradient term $J_{\text{pg}}$ and the action-and-return-independent entropy term $J_{\text{entropy}}$.
We focus on the gradient component from $J_{\text{pg}}$, reconstructed as
\begin{align}
    (\nabla_{b_{\text{last}}}J_{\text{pg}})^{\text{rec}} = \nabla_{b_{\text{last}}} - (\nabla_{b_{\text{last}}}J_{\text{entropy}}(p^{\text{rec}})),
\end{align}
to reconstruct the action and return.
Given the calculation in (\ref{eq: reinforce objective}), the direction of this component depends on the action, while the magnitude depends on the return.
To decide the action, we propose to enumerate all possible actions and calculate the corresponding 
$\nabla_{b_{\text{last}}}J_{\text{pg}}(p^{\text{rec}},a^{\text{rec}},R_t^{\text{rec}})|_{R_t^{\text{rec}}=1}$
by feeding $(p^{\text{rec}},a^{\text{rec}},R_t^{\text{rec}})$ through the original objective and calculating the gradient. 
We select the action that produces the direction of $(\nabla_{b_{\text{last}}}J_{\text{pg}})^{\text{rec}}$.
The return is reconstructed as 
\begin{align}
    R_t^{\text{rec}} = \frac{(\nabla_{b_{\text{last}}}J_{\text{pg}})^{\text{rec}}}{\nabla_{b_{\text{last}}}J_{\text{pg}}(p^{\text{rec}},a^{\text{rec}},\tilde{R}_t)}\Big|_{\tilde{R}_t=1},
\end{align}
where $\tilde{R}_t$ is a constant 1.
This method requires the last bias parameter and an accurate reconstruction of the probability. Otherwise, the solution may align with optimization-based gradient inversion.
When adopting RGI to attack other algorithms, 
the specific action-dependent and return-dependent gradient components should be used for reconstruction.

%% file: experiment.tex
\begin{table*}[t]
    \centering
    \caption{PSNR $(\uparrow)$ and SSIM $(\uparrow)$ of the RGB and depth images.}
    \scalebox{1}{
    \begin{tabular}{ccc|c|cccccccc}
        \toprule
        method & setting & State & Metric & bathroom & bedroom & kitchen & living room & average & best  \\
        \midrule
        \multirow{4}{*}{QGI} & \multirow{4}{*}{S1} & \multirow{2}{*}{RGB} & PSNR & $22.81\pm5.99$ & $19.73\pm5.16$ & $21.41\pm5.69$ & $20.11\pm6.17$ & $21.02\pm5.86$ & 32.17\\
        & & & SSIM & $0.761\pm0.141$ & $0.735\pm0.146$ & $0.724\pm0.167$ & $0.696\pm0.170$ & $0.729\pm0.157$ & 0.929 \\
        & & \multirow{2}{*}{depth} & PSNR & $24.89\pm10.43$ & $22.85\pm7.42$ & $23.26\pm7.17$ & $21.72\pm7.05$ & $23.18\pm8.17$ & 38.09 \\
        & & & SSIM & $0.809\pm0.249$ & $0.823\pm0.161$ & $0.803\pm0.197$ & $0.793\pm0.162$ & $0.807\pm0.195$ & 0.983 \\
        \midrule
        \multirow{2}{*}{QGI} & \multirow{2}{*}{S2} & \multirow{2}{*}{depth} & PSNR & $29.78\pm7.33$ & $26.14\pm7.67$ & $26.45\pm7.27$ & $23.49\pm7.29$ & $26.46\pm7.68$ & 45.76 \\
        & & & SSIM & $0.930\pm0.111$ & $0.899\pm0.152$ &$0.918\pm0.092$ & $0.894\pm0.129$ & $0.910\pm0.123$ & 0.994 \\
        \midrule
        \multirow{4}{*}{RGI} & \multirow{4}{*}{S1} & \multirow{2}{*}{RGB} & PSNR & $18.03\pm5.76$ & $16.98\pm5.93$ & $17.65\pm4.85$ & $15.84\pm4.60$ & $17.12\pm5.35$ & 31.20 \\
        & & & SSIM & $0.564\pm0.190$ & $0.550\pm0.186$ & $0.551\pm0.166$ & $0.511\pm0.163$ & $0.544\pm0.176$ & 0.924 \\
        & & \multirow{2}{*}{depth} & PSNR & $19.75\pm6.66$ & $18.80\pm6.26$ & $18.71\pm5.35$ & $17.04\pm5.17$ & $18.58\pm5.94$ & 36.41 \\
        & & & SSIM & $0.582\pm0.197$ & $0.590\pm0.164$ & $0.594\pm0.158$ & $0.559\pm0.141$ & $0.581\pm0.166$ & 0.940 \\
        \midrule
        \multirow{2}{*}{RGI} & \multirow{2}{*}{S2} & \multirow{2}{*}{depth} & PSNR & $39.19\pm5.34$ & $37.87\pm3.57$ & $36.91\pm4.31$ & $34.85\pm4.99$ & $37.21\pm4.84$ & $58.83$ \\
        & & & SSIM & $0.972\pm0.018$ & $0.968\pm0.018$ & $0.963\pm0.022$ & $0.953\pm0.024$ & $0.964\pm0.022$ & $1.000$ \\
        \bottomrule
    \end{tabular}}
    \label{tab:psnr-ssim}
    \vspace{-2mm}
\end{table*}

\begin{table*}[t]
    \centering
    \caption{Ablation study for error-direction-based method (Opt: joint optimization, Rule: rule-based, Cos: cosine similarity).}
    \scalebox{1.}{
    \begin{tabular}{c|c|ccc|ccc}
        \toprule
        Method & Gradient Calculation & Action & Target Q & Loss & IoU $(\uparrow)$ & $\epsilon(\hat{Q})$(\%) $(\downarrow)$ & $\epsilon(Q)$(\%) $(\downarrow)$ \\
        \midrule
        based on DLG~\cite{zhu2019deepleakage} & objective $J$ & Opt & Opt & MSE & $0.649\pm0.401$ & $1083.64\pm6434.83$ & $89.85\pm12.28$ \\
        based on DLG~\cite{zhu2019deepleakage} & objective $J$ & Rule & Opt & MSE & $0.610\pm0.408$ & $1112.33\pm6676.23$ & $92.44\pm8.42$ \\
        based on inverting~\cite{geiping2020inverting}  & objective $J$ & Rule & Opt & Cos & $0.216\pm0.342$ & $15751.95\pm93392.97$ & $198.83\pm292.31$ \\
        QGI (ours)  & error direction $\nabla_{b_{\text{last}}}$ & Rule & Rule & Cos & $\textbf{0.912}\pm\textbf{00.192}$ & $\textbf{1.03}\pm\textbf{01.59}$ & $\textbf{0.67}\pm\textbf{01.47}$ \\
        \bottomrule
    \end{tabular}}
    \label{tab: QGI vs DLG}
    \vspace{-3mm}
\end{table*}

\textbf{Experimental setup.}
We conducted evaluations of our method on an active perception task using the AI2THOR simulator \cite{kolve2017ai2thor}. In the primary setting (S1), private observations were collected using an RGB and a depth camera, each 
with a resolution of $150\times150$. Additionally, we explored a setting where only a depth camera was available (S2), aiming to examine the privacy risks associated with widely used depth images in robotic tasks.
The target object's bounding box was specified by the 4-dimensional coordinate from the upstream task within the range of $[-1,1]$. 
Gradients were calculated for individual data samples fed into an initialized network. 
$\lambda$ is 0.1.
Optimization-based reconstruction involved $2\times10^4$ iterations with 240 pre-collected samples.
More settings and results can be found on our \href{https://rl-privacy.github.io/RL-gradient-inversion/}{website}.

\textbf{Baselines.} DLG~\cite{zhu2019deepleakage} and inverting gradient~\cite{geiping2020inverting} use optimization-based gradient inversion to reconstruct the image input of the classification task. DLG~\cite{zhu2019deepleakage} uses MSE as gradient matching loss while the inverting gradient~\cite{geiping2020inverting} uses cosine similarity loss. 
Both calculate the reconstructed gradient from the original objective function. In contrast, QGI and RGI leverage the surrogate objective function. 
DLG~\cite{zhu2019deepleakage} optimizes the supervisory signal reconstruction jointly with the input data, while Inverting gradient~\cite{geiping2020inverting} adopts the rule-based method of~\cite{zhao2020idlg} when batch size is 1. 
The direct application of these methods led to low image reconstruction accuracy in our setting. Thus, our comparison with these methods is limited to vector and scalar data in the ablation study section with DQN as the victim algorithm.

\textbf{Metric.}
To evaluate the quality and accuracy of the reconstructed state, action, predicted Q-value, target Q-value, and return, we have selected several metrics for evaluation.
For the image state, we use \textbf{peak-signal-to-noise ratio (PSNR)} and \textbf{structure-similarity index measure (SSIM)} \cite{ssim}. Both are computed on the Y channel of the YCbCr representation for RGB images and the single channel for depth images. For the AI2THOR environment, we normalize the reconstructed RGB images so that the brightest pixel has a value of 255 in the S1 setting, and the corresponding depth image is scaled by the same factor. 
For the vector state (bounding box), we calculate the \textbf{intersection over union (IoU)}. 
The accuracy of action is evaluated by counting the number of accurate results.
For the predicted and target Q-values, we employ the \textbf{percentage error}, denoted as $\epsilon(x)=100\times \frac{|x^{rec} - x|}{|x|}$. We included predicted Q-values since they are correlated to target Q-values.
Lastly, we use absolute error to for returns.

\subsection{Quantitative Results}

\textbf{State: RGB and Depth Image.} 
Table~\ref{tab:psnr-ssim} presents the PSNR and SSIM for the images. In the S1 setting, the mean PSNR of RGB images of QGI is greater than 21dB and the mean SSIM is greater than 0.7. This indicates that the images are recognizable to humans and that they can potentially leak information about the layout of the environments and private elements to adversaries. Depth images exhibit higher PSNR and SSIM, suggesting that the sizes of rooms and the distance to objects are at risk of privacy leakage.
In the S2 setting, despite the lack of color information of adversaries, the reconstructed depth images are more accurate. As shown in Table \ref{tab:psnr-ssim}, PSNR and SSIM increase for QGI and RGI in all rooms, primarily due to a lower dimensionality of the data.

\textbf{State: Coordinate.} 
The mean IoU of both QGI and RGI reaches over 0.9 for both settings.
For QGI, the average IoU is shown in Table~\ref{tab: QGI vs DLG}. 
174 samples achieved IoU larger than 0.9999 in the S1 setting, while 217 samples achieved that in the S2 setting. 
For RGI, 238 samples achieved that in both settings.
Despite the large errors of the failure cases, these results indicate that the vector state can be reconstructed.

\textbf{Predicted Q-value, Target Q-value, and Return.}
The percentage errors of predicted and target Q-values are significantly small, with a mean of less than $1.1\%$ and a standard derivative of less than $1.6\%$ for both settings.
The absolute error of the return reconstruction in RGI has a mean value smaller than $1\times10^{-6}$ for both settings.
These results indicate that the supervision signal is at risk of privacy leakage.

\textbf{Action.}
The action identified from the gradient of a single sample is $100\%$ correct in all 240 samples, demonstrating the effectiveness of the rule-based method of both QGI and RGI. 

\begin{figure}[t]
    \centering
    \includegraphics[width=0.48\textwidth]{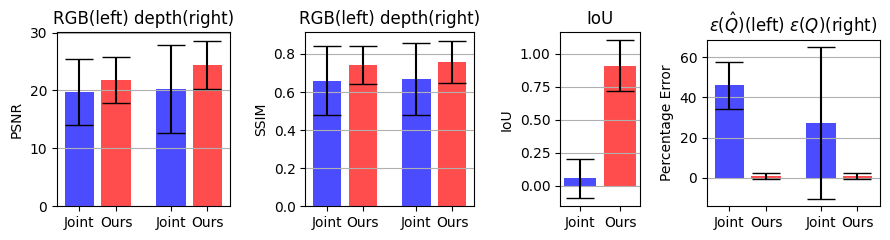}
    \caption{Ablation study of multimodal state reconstruction. Comparison of joint optimization (joint) and our method.}
    \label{fig: ablation joint vs qgi}
    \vspace{-5mm}
\end{figure}

\subsection{Qualitative Results}
The qualitative results are shown in Fig.~\ref{fig:qualitative} for both QGI and RGI in 2 settings.
Results of the S1 setting are shown in the left part. Both the RGB and depth images are reconstructed with recognizable patterns and magnitudes. The RGB images exhibit accurate color tones, despite some colored noise in local areas. 
The depth images 
display similar noisy patterns, but the overall brightness of continuous areas remains accurate. Consequently, the sizes of private rooms can be accessed by adversaries.
In the S2 setting, as depicted in the right part, the reconstructed depth images exhibit accuracy with clear edges and details.

\subsection{Ablation Study}
We evaluate the effect of the proposed separate optimization of the multimodal state by comparing it with the joint optimization of the image state (RGB+depth) and vector state (goal coordinate) in QGI.
As shown in Fig. \ref{fig: ablation joint vs qgi}, the joint optimization shows performance drops for all metrics. 
Besides the lower mean value, PSNR and SSIM have larger standard derivatives, indicating an unstable reconstruction.
The mean IoU is below 0.1, while QGI achieves over 0.9.
Q-value reconstructions also show a performance decline.

We evaluate the proposed error-direction-based gradient calculation by comparing it with the normal gradient calculation from the original objective function. 
As the objective function requires the action and target Q-value as input, these data need to be either reconstructed before the optimization-based gradient inversion or jointly optimized.
We optimize the target Q-value reconstruction jointly with the input and test both the rule-based and jointly optimized action reconstruction. 
As the magnitude of the gradient is essential for the target Q-value reconstruction, we also introduce the MSE gradient matching loss. 
These designs degrade our method to DLG~\cite{zhu2019deepleakage} and inverting gradient~\cite{geiping2020inverting}. As shown in Table~\ref{tab: QGI vs DLG}, our method outperforms these settings. 

%% file: conclusion.tex
This paper investigates the privacy leakage problem in two reinforcement learning algorithms, DQN and REINFORCE, as privacy leakage is a critical concern, but has not received significant attention in the existing reinforcement learning literature.
Given the challenge of adapting existing gradient inversion algorithms to handle multi-modal inputs in RL, we develop a comprehensive pipeline that iteratively reconstructs the inputs. Using experimental results in active perception tasks, we demonstrate that our methods (QGI and RGI) successfully reconstruct the image state, vector state, action, and reward or return from the gradient. Ablation studies further validate the advantage of our proposed design. 
This work is limited as the batch size is restricted to a small number to maintain the quality of the reconstruction and we only applied the attack to two classic RL algorithms. 
We hope to see further research in this area in the future.